\documentclass{article}
\usepackage{kr}

\usepackage{times}
\usepackage{latexsym}

\usepackage{amsfonts}
\usepackage{graphicx}
\usepackage{subfigure}
\usepackage{xspace}
\usepackage{booktabs}   
\usepackage{hyperref}
\usepackage{multirow}   
\usepackage{soul}       
\usepackage{xcolor}
\usepackage{caption}
\usepackage{colortbl}   
\usepackage{pifont}
\usepackage{enumitem}
\usepackage{amsmath}
\usepackage{setspace}
\usepackage{csquotes}
\usepackage{adjustbox}
\usepackage[mathscr]{euscript}

\usepackage{float}

\setul{1.6pt}{.8pt} 

\newcommand{\kgtrsf}{\textsl{\textsc{kg-trsf}}\xspace}

\newcommand{\starbert}{\textsl{\textsc{star}}\xspace}
\newcommand{\lpbert}{\textsl{\textsc{lp-bert}}\xspace}
\newcommand{\drgi}{\textsl{\textsc{drgi}}\xspace}
\newcommand{\boxe}{\textsl{\textsc{boxe}}\xspace}
\newcommand{\kgbert}{\textsl{\textsc{kg-bert}}\xspace}
\newcommand{\bert}{\textsl{\textsc{bert}}\xspace}
\newcommand{\wnrr}{\textsl{\textsc{wn18rr}}\xspace}
\newcommand{\fb}{\textsl{\textsc{fb15k-237}}\xspace}
\newcommand{\jf}{\textsl{\textsc{jf17k}}\xspace}
\newcommand{\spa}{\textsl{\textsc{sp}}\xspace}
\newcommand{\ipa}{\textsl{\textsc{ip}}\xspace}
\newcommand{\hn}{\textsl{\textsc{khn}}\xspace}
\newcommand{\lcc}{\textsl{\textsc{lcc}}\xspace}
\newcommand{\iva}{\textsl{\textsc{iva}}\xspace}
\newcommand{\all}{\textsl{\textsc{all}}\xspace}

\newcommand{\mask}{\mbox{\texttt{[mask]}}}
\newcommand{\nopath}{\mbox{[no\_path]}}
\newcommand{\set}[1]{\ensuremath{\mathcal{#1}}}
\newcommand{\relset}[0]{{\set{R}}}
\newcommand{\entset}[0]{{\set{E}}}
\newcommand{\kg}[0]{{\set{G}}}

\usepackage[ruled,vlined,linesnumbered,boxed]{algorithm2e}
\SetAlCapNameFnt{\small}
\SetAlCapFnt{\small}
\SetAlgorithmName{Algo}{Algo}{List of Algorithms}
\SetKwInput{KwInput}{Input}
\SetKwInput{KwOutput}{Output}
\SetKwInput{KwReturn}{Return}

\SetCommentSty{mycommfont}
\SetKwComment{Comment}{$\triangleright$\ }{}
\title{Using Graph Algorithms to Pretrain Graph Completion Transformers}

\author{Jonathan Pilault$^{1}$\footnote{ Work performed at ServiceNow Research.} , Michael Galkin$^{1}$, Bahare Fatemi$^{2*}$, \\
\textbf{Perouz Taslakian}$^{3*}$, \textbf{David Vasquez}$^{4}$, \textbf{Christopher Pal}$^{1,4,5}$ \\
\affiliations
$^1$Polytechnique Montreal \& Mila, $^2$ Google Research, \\
$^3$Samsung AI, $^4$ServiceNow Research, $^5$Canada CIFAR AI Chair \\
\emails\{jonathan.pilault,michael.galkin,christopher.pal\}@mila.quebec
}

\begin{document}

\maketitle

\begin{abstract}
Recent work on Graph Neural Networks has demonstrated that self-supervised pretraining can further enhance performance on downstream graph, link, and node classification tasks. 
However, the efficacy of pretraining tasks has not been fully investigated for downstream large knowledge graph completion tasks.
Using a contextualized knowledge graph embedding approach, we investigate five different pretraining signals, constructed using several graph algorithms and \emph{no external data}, as well as their combination.
We leverage the versatility of our Transformer-based model to explore \emph{graph structure generation} pretraining tasks (i.e. path and k-hop neighborhood generation), typically inapplicable to most graph embedding methods.
We further propose a new path-finding algorithm guided by information gain and find that it is the best-performing pretraining task across three downstream knowledge graph completion datasets.
While using our new path-finding algorithm as a pretraining signal provides 2-3\% MRR improvements, we show that pretraining on all signals together gives the best knowledge graph completion results.
In a multitask setting that combines all pretraining tasks, our method surpasses the latest and strong performing knowledge graph embedding methods on all metrics for \fb, on MRR and Hit@1 for \wnrr and on MRR and hit@10 for \jf (a knowledge hypergraph dataset).

\end{abstract}

\section{Introduction}
\label{sec:intro}


Transfer learning has emerged as a powerful technique in several domains \cite{s2s-unsupervised,glove,tomar-etal-2017-neural,subramanian2018learning,conv_transf2014,yosinski2014transferable,selfie2019}, in which a model is first pretrained on relevant tasks before being fine-tuned on a downstream task.
In most modern vision and NLP applications, such pretraining is often based on the versatile Transformer model \cite{transformer2017} using self-supervised learning on unlabeled data \cite{bert,t5,bao2022beit}.
Transformer-based and self-supervised pretraining have also been applied in the graph representation learning scenarios. 
These studies, however, focus mostly on non-relational graphs \cite{bgrl,slaps} or small molecular graphs \cite{graphormer},
leaving pretraining approaches for Knowledge Graphs (KG) link prediction relatively unexplored.

Many Graph Neural Network (GNN) techniques use positional embeddings of the entities in the relation along with entity/relation embeddings to represent tuples \cite{you2019position}.
Transformer Language Models \cite{transformer2017} use positional embeddings in a similar way. 
Transformers can perform \emph{contextualized} link prediction by masking out one of the entities in an input tuple.
In an arity-2 graph, a KG can be viewed as a set of triplets $\mathcal{G} = \{(h_i, r_i, t_i)\}$, $i\in \{1,\dotsc, T\}$, where $h_i$ is the head entity, $t_i$ is the tail entity, and $r$ is the relation between $h_i$ and $t_i$. 
We can feed the sequence $h_i$, $r_i$, \mask{} or \mask{}, $r_i$, $t_i$, where \mask{} represents the  entity to predict.
Since Transformers can process sequences of arbitrary length, we can go beyond triple-based KGs.
We show that our approach can also be applied to the more general case of  \emph{Knowledge Hypergraphs} (KHGs,~\citeauthor{fatemiijcai2020}), where edges are composed of tuples with an arbitrary number of entities.
It would require many modifications to apply KHGs to other KG Embedding methods~\cite{fatemiijcai2020}.
With the versatility of our base model, we can also perform other tasks, such as autoregressive path generation or query conditioned neighborhood prediction, that GNNs or KG embeddings cannot.

So far, the most common pretraining approach to KG link prediction is to use a Transformer pretrained on a large corpora of text, such as Wikipedia \cite{Yao2019KGBERTBF,star2021,lpbert2022}.
There are two issues with using pretrained language models:
(1) Knowledge or information leakage may occur from the pretraining text data to common KG completion benchmarks such as \fb \cite{freebase}, extracted from Wikidata (formerly Freebase) and \wnrr \cite{wn18rr}, extracted from Wordnet~\cite{miller-1994-wordnet};
(2) If the pretraining text corpora is out of domain with respect to the test KG, it is unclear if external data unrelated to the targeted knowledge base will be helpful for knowledge completion.
Further, not using external data is instrumental in situations where data is scarce, confidential, or classified such as medical records, crime networks, or robotics~\cite{Wang2019/07} and it may be infeasible to pretrain a Transformer on a more closely related domain of knowledge.
Non-parametric link prediction based on graph structure (e.g.: Katz index, Adamic Adar) have been known to do well on homogenous graphs but not on heterogeneous graphs~\cite{path_link_pred_2020}. 
We posit that pretraining with a Transformer encoder that learns similar structural heuristics can also be applied to heterogeneous graphs. 
Pretraining has shown to have better generalization for various biology/chemistry datasets \cite{Hu2020Strategies} and we show that it is also true for knowledge graphs.

In this paper, we study the effectiveness of various graph-based pretraining schemes, that \emph{do not use external data}, applied to a Transformer model, trained either separately or jointly, and evaluated on the downstream task of KG completion (link prediction).
We look into various graph structural signals, such as paths between two nodes, local neighborhoods, adjacency matrices permutations or clustering coefficients, that are derived from common graph-algorithms.
To the best of our knowledge, we are the first KG completion work that investigates multiple self-supervised pretraining signals based on the graph structure.
We introduce a new path-finding algorithm after noticing that:
(1) Encoding random paths has proven beneficial in several graph reasoning and completion methods~\cite{Mazumder2017ContextawarePR,pmlr-v77-jiang17a,wang-etal-2020-connecting,xu-etal-2021-p-int};
(2) Random paths that represent a reasoning chain do not always make logical sense (see Section~\ref{sec:info_path}).
Our techniques can also be applied to a generalization of KGs, i.e. KHGs.

Our contributions are as follows:
\begin{enumerate}
\item We report results of five pretraining objectives suitable for our Transformer-based contextualized KG embedding method in Section~\ref{sec:method}; 
\item We introduce a new path-finding algorithm (\ipa) that is guided by the Kullback-Leibler divergence (i.e. information gain) in Section~\ref{sec:info_path};
\item We evaluate the effect of pretraining strategies on triple-based KGs as well as on KHGs and show that our new path-based (\ipa) and our multitask (\all) pretraining schemes perform best for both low and high density knowledge graphs in Section~\ref{sec:exp}.
 \end{enumerate}

\section{Related Works}
\label{sec:lit_review}

Our method is most related to \emph{Contextualized KG Embeddings}\cite{coke} and \emph{Self-Supervised Graph Representation} pretraining. We will discuss the two lines of work below.

\paragraph{Knowledge Graph Embedding} methods encode distributed representations of entities and relations using a continuous vector. 
The representation is learned via a scoring function that measures the plausibility that a given triple $\{(h_i, r_i, t_i)\}$ exists. Such embeddings maximize the plausibility of observed triples, which may not be predictive enough for downstream tasks \cite{kg-emb-rules-2015,large-kgcomp-2015}. 
To make the embeddings more transferable, researchers have incorporated other types of information. 
For example, \emph{external} data sources can complement a KG embedding by also encoding entity types \cite{guo-etal-2015-semantically,hiear-kgemb-2016} or textual descriptions \cite{kgemb_descrip-2016,text-enhanced-kgemb-2016}.
As explored here, we can also enhance KG embedding by \emph{only} using structural features from the graph. For example, relational paths and multi-hop relationships between entities (see section \ref{app:def} for more details), are a type of structural information that has proven useful \cite{lin-etal-2015-modeling,toutanova-etal-2016-compositional,das-etal-2017-chains} for KG completion. 

\paragraph{Self-Supervised Graph Representation} learning techniques broadly fall into three categories and include methods that: 
(1) use random walk procedures to encode diverse neighborhoods \cite{deepwalk2014,grover2016node2vec,NIPS2017_5dd9db5e}; 
(2) reconstruct a graph's adjacency matrix \cite{vgae2016,NEURIPS2019_fd4771e8,slaps}; 
(3) maximize mutual information between local node representations and global graph representations \cite{velickovic2018deep,NEURIPS2020_3fe23034,drgi2021}. 
For applications in chemistry and biology, most pre-training techniques also use additional datasets and signals like MAE loss \cite{Hu_2020Strategies,graphormer}, however, our technique does not use \emph{external} data. 
With the exception of Deep Relational Graph Infomax (\drgi)~\cite{drgi2021}, most self-supervised pretraining schemes based on graph structure have never been applied to solve large KG completion tasks. 
Other techniques such as \kgbert \cite{Yao2019KGBERTBF} use a \bert-based \cite{bert} pretrained language models with entity descriptions for KG link prediction. 
Again, such work relies on external data and additional information (descriptions) while our pretraining signals are self-contained.

\section{Definition and Notation}
\label{app:def}

In this section, we provide definitions and background on link prediction, paths, multi-hop neighborhoods and graph adjacency matrices.
We also introduce important notations used in equations and algorithms.

Since we are also evaluating our technique on the KHG \jf, a generalization of KG with arity-$n$, we provide in this section definitions based on hypergraphs. 
Without loss of generality, in this section, we formulate our definitions and notation based on hypergraphs. 
Given a finite set of entities $\entset$ and a finite set of relations $\relset$, a
\emph{tuple} is an ordered set of the form $r(e_1, \dots, e_n)$, where $r \in \relset$, $e_i \in \entset$ for all $i=1,\dots,n$ and $|r| = n$ is its \emph{arity}. 
Let $\kg$ be the set of ground truth tuples; that is, it specifies all of the tuples that are true so that if a tuple is not in~\kg, it is false.
A \emph{knowledge hypergraph} (KHG) consists of a subset of the tuples $\mathcal{G'} \subseteq \kg$.
A knowledge graph is a special case of a knowledge hypergraph where all relations have arity $2$. 
We let $E(r) \subseteq \entset$ denote the set of all entities that appear in a tuple in $\kg'$ having relation $r$. 
That is, $E(r) = \{e_i | r(\dots, e_i, \dots) \in \kg'\}$.
Similarly, we let $R(e_i) \subseteq \relset$ denote the set of all relations that appear in a tuple in $\kg'$ linked to entity $e_i$. 
That is, $R(e_i) = \{r | r(\dots, e_i, \dots) \in \kg'\}$.
We say that two tuples are \emph{incident} in $\kg'$ if they share at least one entity. 
Two entities are \emph{connected} in \kg' if they appear together in a tuple.
For example, $r_1(e_1, e_2, e_3)$ is incident to $r_2(e_4, e_3, e_5)$ because they share $e_3$. 
Entities  $e_2$ and $e_3$ are connected because they both appear in the first tuple. 
A query for the knowledge completion task is a tuple with one missing entity that needs to be predicted. 
We let $Q_r = [r, e_1, \dots, \mask, \dots, e_n]_q$ denote a query tuple with relation $r$, where $\mask{}$ is the placeholder token (the masked-out entity) for the entity we want to predict.

A \emph{path} $P$ in a KHG is a sequence of tuples where two consecutive tuples share at least one entity.  
We say that $P$ \emph{connects} entities $e_i$ and $e_j$ if the first tuple in $P$ contains $e_i$ and the last tuple contains $e_j$. 
A \emph{relational path} $P_{rel}$ between $e_i$ and $e_j$ is the sequence of relations along the edges of a path connecting $e_i$ and $e_j$ such that no relation is repeated along the path (without cycles).
For example, $e_1$ and $e_6$ are connected through path $P: r_1(e_1,e_2, e_3), r_2(e_3,e_4, e_5), r_3(e_4,e_5, e_6)$, and have a relational path $P_{rel}: r_1, r_2, r_3$. Given the set of entities in queries $\{Q_r\}_r$, we define the set of all possible paths between pairwise entities as $\{P_{rel}\}_{r}$.

The \emph{k-hop neighborhood} of entity $e_i$, denoted $\mathcal{N}_k(e_i)$, is the unordered set of entities $E_{\mathcal{N}_k} =  \{e_k, \dots, e_j\} \subseteq \entset$ enclosed in a $k$-hop radius around $e_i$. 
We denote the relation-less adjacency matrix of entities in $\mathcal{N}_k(e_i)$ as 
$\mathbf{A}_{E_{\mathcal{N}_k}}\in 
\mathbb{R}^{|E_{\mathcal{N}_k}| 
\times |E_{\mathcal{N}_k}|}$. 
For a relational graph $\mathcal{G'}$, we define a relation-less adjacency matrix $\mathbf{A}_{E}$ as:
\begin{align}
\mathbf{A}_{E} = \sum_{r \in \relset} \mathbf{A}_{E(r)}, \mathbf{A}_{E} \in \mathbb{R}^{|E_{\entset}| \times |E_{\entset}|}.
\label{eq:adj}
\end{align}
Adjacency $\mathbf{A}$ has a certain order of appearance of entities in columns and rows which is arbitrarily set by the order in which dataset tuples are sampled. 
We let $\widetilde{\mathbf{A}}$ be the equivalent adjacency matrix with a permuted order of entities in columns and rows (e.g.: $r_1, r_2, r_3, r_4$ can become $r_2, r_3, r_1, r_4$). 
The local clustering coefficient $c_{e_i}$ \cite{Watts-Colective-1998} of $\mathcal{N}_k(e_i)$ measures the proportion of closed triangles in the local k-hop neighborhood of an entity such that:
\begin{align}
c_{e_i} = \frac{|(e_1,e_2\in \entset : e_1,e_2 \in \mathcal{N}_k(e_i)|}{\binom{d_{e_i}}{2}},
\label{eq:lcc}
\end{align}
where $d_{e_i}=\sum_{e_j \in \mathcal{N}_k(e_i)}\mathbf{A}[e_i,e_j]$ is the node degree of $e_i$. For the pretraining task in section \ref{sec:3hop}, we define the probability that an entity $e_i$ is found in a in $k$-hop neighborhood (probability of occurrence) $O(e_i)$ as:

\vspace{-0.6cm}
\begin{align}
O(e_i) = \frac{d_{e_i}}{\sum_{e_j \in E_{\mathcal{N}_k}}d_{e_j}}, \text{ and } \mathbf{O}_{\mathcal{N}_k} = 
\begin{bmatrix}
O(e_1) \\
\vdots \\
O(e_{|\entset|})
\end{bmatrix}
\label{eq:occurence}
\end{align}

\section{Methodology}
\label{sec:method}

In this section, we provide the motivation and details of the five different graph algorithms that we employed to create pretraining tasks for our experiments: 
(1) Relational \underline{S}hortest \underline{P}ath sequence generation (\textbf{\spa}); 
(2) \underline{I}nformation gain \underline{P}ath sequence generation (\textbf{\ipa}); 
(3) \underline{K}-\underline{H}op \underline{N}eighbor prediction (\textbf{\hn}); 
(4) \underline{I}n\underline{v}ariant \underline{A}djacency matrix classification (\textbf{\iva}); 
(5) \underline{L}ocal \underline{C}lustering \underline{C}oefficient estimation (\textbf{\lcc}). 
Please see Section \ref{app:def} for more details on definitions and notation used in this section.
When jointly trained (\textbf{\all}), we use all pretraining tasks and prepend a task token to delineate each task.
Note that for \all, the total is the unweighted sum of each pretraining task's loss.
An overview of the task objectives is outlined in Table \ref{tab:pretrain_task}.

\begin{table}[th]
\small
\setlength\tabcolsep{2.5pt}
\begin{center}
\begin{tabular}{llllcc}
\toprule
Task & Input & Target & Objective & Type \\
\midrule
\rowcolor{cyan!16}\multicolumn{5}{c}{\it \textbf{Path-based}} \\
\spa & \multirow{2}{*}{$\{e_i,r,e_j\}$} & \multirow{2}{*}{$\{P_{rel}\}_{r}$} & \multirow{2}{*}{$\prod_{k}^{n} f(p_k|p_{<k}, e_i, r, e_j)$} & \multirow{2}{*}{SG} \\
\ipa & & & &  \\
\rowcolor{cyan!16}\multicolumn{5}{c}{\it \textbf{Neighborhood Based}}  \\
\hn  & $\{e_i,E_{\mathcal{N}_{k-1}}\}$ & $\mathbf{O}_{\mathcal{N}_k}$ & $\text{KL}(\mathbf{O}_{\mathcal{N}_k}||f(e_i,E_{\mathcal{N}_{k-1}}))$ & MP \\
\iva & $\{\mathbf{A}, \widetilde{\mathbf{A}}|\mathbf{A}'\}$ & $\{1|0\}$ & $y_i log(f(\mathbf{A},\widetilde{\mathbf{A}}|\mathbf{A}'))$ & BC \\
\lcc & $\{e_i,E_{\mathcal{N}_{k-1}}\}$ & $c_{e_i}$ & $(f(e_i) - c_{e_i})_k^2$ & R\\
\bottomrule
\end{tabular}
\caption{\label{tab:pretrain_task} Overview of Graph Algorithm Pretraining Tasks. Rel=Uses Relations; SG=Sequence Generation; BC=Binary Classification; MP=Multi-label Prediction; R=Regression.}
\end{center}
\end{table}



\subsection{Relational Shortest Paths (\spa)}
\label{sec:shortest_path}

Using the definition of global and quasi-local graph structural information from \cite{Benhidour2022AnAF}, path signals capture global graph structure since all paths cross each node in the graph at least once. 
The idea to use paths is loosely inspired by global methods such as the Katz index that considers sets of paths between two query nodes to measure the likelihood of a link between them \cite{Katz1953ANS}.

The set of relational paths $\{P_{rel}\}_{r}$ as defined in Section~\ref{app:def} may yield an exponential number of choices. 
In practice, we would like to limit the number of such paths and hence, we need a way to select a subset of $\{P_{rel}\}_{r}$.
There are a few possible heuristics for selecting a subset of such paths. 
For example, we can find shortest paths based on Dijkstra's algorithm and then just keep the sequences of relations joining two entities $\{e_i, r, e_j\}$. 
As specified in Table \ref{tab:pretrain_task}, we condition our path sequence generation with $\{e_i, r, e_j\}$.
The next token generated is the first relation on a sampled path $\{P_{rel}\}_{r}$.
Our path-based pretraining algorithms allow the model to explore $\kg$ beyond the tuples included in $\mathcal{G'}$ since $\kg$ includes relational paths between entities that may not appear in any of the train, validation, or test sets.
When no path exist between $\{e_i,e_j\}$, the model target is the \nopath{} token.
We posit that the quality of the paths is also important and created a new path finding algorithm called \ipa to test this hypothesis.

\subsection{Information gain Paths (\ipa)}
\label{sec:info_path}

\begin{algorithm}
\small
\caption{\small Top\_k Path Information Gain}
    \SetAlgoLined
    \DontPrintSemicolon
    \label{alg:info_gain_rel_path}  
    \SetCustomAlgoRuledWidth{1.2cm}  
    \KwInput{Training set \(\mathcal{G'}\); query relation $r$; $k$ number of top paths; max relational paths length (max hops) $l$}
    \KwOutput{at most $n=k^l$ relational paths $\{P_{rel}\}_r$;}

    \(R' \gets \text{FindIncidentRelations}(r, \emptyset)\) \tcp{Algorithm \ref{al:incident_rel}}
    \(R' \gets \text{TopEntropy}(k, R')\) \tcp{Algorithm \ref{al:topk_entropy}}
    \(\{P_{rel}\}_r \gets R'\) \;
    \For{\(i \in \{1, \dotsc, l\} \)}{
        \For{\( r' \in R' \); \( p \in \{P_{rel}\}_r \)}{
            \( R'' \gets \text{FindIncidentRelations}(r')\) \;
            \(R'' \gets \text{BottomCondEntropy}(k, R'', r')\) \tcp{Algorithm \ref{al:topk_entropy}}

            \For{\( r'' \in R'' \)}{
                \If{$r'$ is the last element in $p$}{
                    \(\{P_{rel}\}_r \gets \{p\}\ \cup \{r''\} \)
                }
            }
        }
        \( R' \gets R'' \)
    }
    \KwReturn{$\{P_{rel}\}_r$}
\end{algorithm}

While \spa provides a measure of subgraph distance and connectivity, i.e. the number of elements between entities $\{e_i,e_j\}$ in the graph, it may often yield reasoning chains having little semantic overlap with the original query. 
For example, a masked query ``[Spain] [form of government] $\mask{}$'' produces many \spa such as ``[nominated for], [film/country]'', which is a sequence of relations unrelated to ``[form of government]''.
Further, \spa algorithms select paths that minimize the distance (number of hops) between two entities. 
It is therefore formed from shortest paths of the same length, which may limit subgraph exploration.
We propose a new path finding algorithm, \emph{Information Gain Paths} in Algorithm \ref{alg:info_gain_rel_path}. 
\ipa has the same initial condition $\{e_i, r, e_j\}$ to start generating the paths and has the same learning objective as \spa.
\ipa also helps us reduce the number of paths by selecting the ones that constitute a beneficial context for answering the query based on a measure of information gain. 
Given a query tuple $Q_r$ (having relation $r$), the algorithm progressively builds relational paths $\{P_{rel}\}_{r}$ starting from a relation $r$ such that, at each step, it selects the (at most) $k$ \emph{incident} relations that would yield the $k$ highest information gain for the paths constructed so far.

\begin{algorithm}
\small
\caption{\label{al:incident_rel}  Finding all relations incident to $r$ in $\kg'$.}
    \SetKwFunction{FMain}{FindIncidentRelations}
    \SetKwProg{Fn}{Function}{:}{}
    \Fn{\FMain{$r$, $P_{rel}$}}{
        \(R' \gets \emptyset\)\;
        \For{\(r' \in \relset \backslash \{r\}\)}{
            \If{\(E(r') \cap E(r) \neq \emptyset \textrm{\textbf{ and }}  r' \notin P_{rel}\)}{
                \(R' \gets r'\) \;
            }
        }
        \textbf{return} $R'$
    }
    \textbf{End Function}\;
\end{algorithm}

\begin{algorithm}
\small
\caption{\label{al:topk_entropy} Finding $r \in R$ with top $k$ entropy $H(r)$ or with with bottom $k$ conditional entropy $H(r|r')$.}
    \SetKwFunction{FMain}{TopEntropy}
    \SetKwProg{Fn}{Function}{:}{}
    \Fn{\FMain{$k$, $R'$}}{
        \(\text{Compute $H(r')$ for all the relations $r' \in R'$ (Eq. \eqref{eq:entropy})}\)
        \(\textbf{return} \text{ set ${r' \in R'}$ with top-k highest $H(r')$}\)
    }
    \textbf{End Function}\;
    \SetKwFunction{FMain}{BottomCondEntropy}
    \SetKwProg{Fn}{Function}{:}{}
    \Fn{\FMain{$k$, $R''$, $r'$}}{
        \(\text{Compute $H(r''|r')$ for all $r'' \in R''$ (Eq. \eqref{eq:conditional_entropy})}\)
        \(\textbf{return} \text{ set ${r''}\in R''$ with k lowest $H(r''|r')$}\)
    }
    \textbf{End Function}\;
\end{algorithm}


With infinite $k$, the algorithm is identical to the breadth-first search as no relations are ignored. Algorithm~\ref{alg:info_gain_rel_path} bears some similarities with beam search \cite{beamsearch}. 
There are however a few differences: (1) for $l$ hops and $k$ beam size, we obtain a maximum of $k^l$ paths; (2) a relation already in $P_{rel}$ cannot be used to form the next hop; (3) the paths are formed via a back-chaining process starting from the last relation on path $P_{rel}$ that connects $e_i$ and $e_j$; (4) an extra step is performed to select paths in $\{P_{rel}\}_{r}$ linking entities in $Q_r$.

We let $\text{IG}({P_{rel}})$ denote the \emph{Information Gain} of relational path $P_{rel}: r_1, \dotsc, r_l \in \kg'$ having length $l$  as $\text{IG}({P_{rel}}) = H(r_l) - \sum\limits_{i=1}^{l-1} H(r_{l-i}|r_{l-i+1})$.
We define the \emph{Information Entropy} $H(r_l)$ for a relation $r_l$, and entities $E(r_l)$ that appear in a tuple in $\kg'$ having relation $r_l$ as:
\begin{equation}
\scriptstyle
\label{eq:entropy}
    H(r_l) =  -\left(\frac{|E(r_l)|}{|\entset|}\log(\frac{|E(r_l)|}{|\entset|}) + \frac{|\entset \setminus E(r_l)|}{|\entset|}\log(\frac{|\entset \setminus E(r_l)|}{|\entset|})\right),
\end{equation}
where $\entset$ is all entities in $\kg'$. 

The \emph{Conditional Information Entropy} $H(r_{i-1}|r_i)$ of two consecutive relations  is defined as:
{\small
\begin{align}\label{eq:conditional_entropy}
    H(r_{i-1}|r_i) =  
    & -\frac{|E(r_{i-1})|}{|\entset|}
    \biggl (\mathrm{U} \log(\mathrm{U})
    + \mathrm{V}\log(\mathrm{V})\biggr)
\end{align}
}

where $\mathrm{U} = \frac{ |E(r_{i-1}) \cup E(r_{i}) \setminus E(r_{i-1}) \cap
    E(r_{i})|}{ |E(r_{i-1})| }$ and $\mathrm{V}=\frac{ |E(r_{i-1}) \cap E(r_{i})| }{ |E(r_{i-1})| }$.

\vspace{10pt}
Algorithm~\ref{alg:info_gain_rel_path} can be summarized in a four step process that bear resemblance with information gain criterion in decision trees: 
\begin{enumerate}
\item Starting a query relation $r$, we first collect all other relations that are incident in the graph with \textit{FindIncidentRelations}.
\item We then choose the next $k$ most probable and informational relations (excluding $r$) given information entropy in Equation~\ref{eq:entropy} and using the \textit{TopEntropy} function.
\item We then choose the next $k$ relations that have the highest information value with respect to relations in the previous step, i.e. that have the \emph{lowest} conditional information entropy from Equation~\ref{eq:conditional_entropy} using the \emph{BottomCondEntropy} function.
\item Repeat step 3 until we reach $l$ hops. 
\end{enumerate}

\begin{table*}[ht!]
    \centering
    \setlength\tabcolsep{3.5pt}
    
    \begin{tabular}[b]{l|c|cccc|cccc||cccc}
        \toprule
        \multirow{2}{*}{} & \textbf{Pre}     & \multicolumn{4}{c|}{\textbf{\fb}} & \multicolumn{4}{c||}{\textbf{\wnrr}} & \multicolumn{4}{c}{\textbf{\jf}} \\
                          & \textbf{Trained} & \textbf{MRR} & \textbf{H@1} & \textbf{H@3} & \textbf{H@10}  
                                                                & \textbf{MRR} & \textbf{H@1} & \textbf{H@3} & \textbf{H@10}       
                                                                & \textbf{MRR} & \textbf{H@1} & \textbf{H@3} & \textbf{H@10}      \\
        \rowcolor{gray!20}\multicolumn{14}{c}{\it \textbf{Contextualized KG Embedding Results}}        \\
        \kgtrsf$^1$ & $\times$   & 0.364 & 0.272 & 0.400 & 0.549                    
                                 & 0.484 & 0.450 & 0.496 & 0.553                                   
                                 & 0.532 & 0.445 & 0.561 & 0.687        \\
        \kgtrsf \spa& \checkmark & 0.368 & 0.275 & 0.405 & 0.561                    
                                 & 0.489 & 0.452 & 0.499 & 0.566                                   
                                 & 0.537 & 0.450 & 0.569 & 0.703        \\
        \kgtrsf \ipa (ours)      & \checkmark & \underline{0.372} & \underline{0.277} & \underline{0.412} & \underline{0.579}                    
                                 & \underline{0.491} & \underline{0.454} & 0.507 & 0.575                                   
                                 & 0.549 & 0.457 & 0.582 & 0.723        \\
        \kgtrsf \hn & \checkmark & 0.365 & 0.273 & 0.401 & 0.550                    
                                 & 0.484 & 0.449 & 0.497 & 0.556                                   
                                 & 0.536 & 0.450 & 0.565 & 0.701        \\
        \kgtrsf \lcc& \checkmark & 0.364 & 0.272 & 0.400 & 0.549                    
                                 & 0.486 & 0.451 & 0.498 & 0.563                                   
                                 & 0.532 & 0.445 & 0.561 & 0.679        \\
        \kgtrsf \iva& \checkmark & 0.365 & 0.272 & 0.401 & 0.555                    
                                 & 0.483 & 0.448 & 0.497 & 0.560                                   
                                 & 0.538 & 0.453 & 0.568 & 0.705        \\
        \kgtrsf \all             & \checkmark & \textbf{0.380} & \textbf{0.279} & \textbf{0.422} & \textbf{0.591}
                                 & \textbf{0.499} & \textbf{0.456} & \underline{0.511} & 0.588
                                 & \textbf{0.554}    & \underline{0.465} & \underline{0.594} & \textbf{0.715} \\
        
        \rowcolor{gray!20}\multicolumn{14}{c}{\it \textbf{SOTA KG Embedding Results}}   \\
        \boxe     & \checkmark          & 0.337 & 0.238 & 0.374 & 0.538                    
                                 & 0.451 & 0.400 & 0.472 & 0.541                                   
                                 & \underline{0.553} & \textbf{0.467} & \textbf{0.596} & \underline{0.711}    \\
        \drgi     & \checkmark   & 0.362 & 0.270 & 0.399 & 0.549                    
                                 & 0.479 & 0.445 & 0.496 & 0.543                                   
                                 & --- & --- & --- & ---    \\
        $\starbert$* & \checkmark   & 0.358 & 0.205 & 0.322 & 0.482                    & 0.401 & 0.243 & 0.491 & \underline{0.675}                                 & --- & --- & --- & ---    \\
        $\lpbert$*   & \checkmark   & 0.310 & 0.223 & 0.336 & 0.490                    
                                 & 0.482 & 0.343 & \textbf{0.563} & \textbf{0.752}                                   
                                 & --- & --- & --- & ---    \\
        \bottomrule
    \end{tabular}
    \caption{
    Link prediction test results \label{tab:link-prediction}. H@=HIT@. Results from: $^1$\protect\citeauthor{coke}. In the table, \textbf{bold} is best and \underline{underline} is second best. \emph{*Uses external data and is a contextualized KG embedding method}. Methods that require too many modifications to work with hypergraphs (\jf) were not evaluated.
    }

\end{table*}

\subsection{k-Hop Neighborhood prediction (\hn)}
\label{sec:3hop}

Our next graph algorithm to create our pretraining data is based on the observation that the representation of nodes in a neighborhood $\mathcal{N}_k(e_i)$ encodes subgraph structure and provides a more powerful representation of $e_i$ \cite{grl_hamilton_2017}. 
\hn is a quasi-local graph structural information since we are trying to predict properties of the graph k-hops away. 
\hn are inspired by structural based similarity indices such as Adamic-Adar~\cite{ADAMIC2003211} and Common Neighbor~\cite{PhysRevE.64.025102}, measures based on both common neighbor and node degree.
To coerce a Transformer model to understand local structure up to $k$ hops, we ask the model to generate $\mathbf{O}_{\mathcal{N}_k}$ of the next hop from Eq. \ref{eq:occurence}, given $E_{\mathcal{N}_{k-1}}$ and $e_i$.
We input an arbitrarily ordered set of entities $E_{\mathcal{N}_{k-1}}$ to condition our prediction.
Note that typically entities are ordered according to their appearance in the original datasets.
The loss to learn output entity occurrence probability of entities for hops up to $k=3$ is $L_{\hn} = \sum^k_{i=1}\big ( \text{KL}(\mathbf{O}_{\mathcal{N}_k}||f(e_i,E_{\mathcal{N}_{k-1}})) \big )$, where $E_{\mathcal{N}_{0}} = \{\}$ and $\text{KL}$ is the Kullback–Leibler divergence \cite{d_kl}.
Please note that we directly chose $k=3$ since it typically performs better than $k=2$ in a variety of settings \cite{nikolentzos2020k}.

\subsection{Invariant Adjacency matrix (\iva)}
\label{sec:iva}
One of the key issues with previous heuristics in Section \ref{sec:3hop} is that a certain order is assumed.
However, graph representations are more powerful when they are order invariant \cite{inv_gnn_2017}. 
To allow order invariance, we propose a binary classification task based on adjacency matrices $E_{\mathcal{N}_k}$ of the local neighborhood subgraph. 
The label is $1$ if the Transformer is given inputs $\{\mathbf{A}, \widetilde{\mathbf{A}}\}$, where $\widetilde{\mathbf{A}}$ is the equivalent adjacency matrix to $\mathbf{A}$ where the entity order is randomly permuted.
The label is $0$ if the Transformer is given inputs $\{\mathbf{A}, \mathbf{A}'\}$, where $\mathbf{A}'$ is  a corrupted adjacency matrix.
We corrupt the adjacency matrix by randomly swapping the columns or by randomly assigning different adjacency matrix values.
Since $\mathbf{A}$ is a symmetric matrix, the input to the Transformer is a flattened sequence of the upper triangular matrix that contains (1) column entities and (2) $\text{row}_1(\mathbf{A}), \dots, \text{row}_{|E_{\mathcal{N}_k}|}(\mathbf{A})$. 
For example, $\mathbf{A} = \begin{bmatrix}
  1 & 2 & 1 \\
  2 & 0 & 3 \\
  1 & 3 & 1 \\
\end{bmatrix}$ with columns $[e_1,  e_2, e_3]$, will produce the flattened sequence $[e_1,  e_2, e_3, 1 , 2, 1, 0, 3, 1]$.
\iva injects order-invariance in the representations of entities and relations. 
For link prediction, order-invariance is useful when aggregating information on a neighborhood.

\subsection{Local Clustering Coefficient (\lcc)}
\label{sec:lcc}

The clustering coefficient is a measure of the degree to which nodes in a graph tend to cluster together.
For \lcc, we use the same inputs as with \hn in section \ref{sec:3hop}.
\lcc is quasi-local graph structural information since we are trying to predict properties of the graph k-hops away.
\lcc has a similar motivation than \hn and is related to Adamic-Adar and Common Neighbor.
However, we are trying to estimate the local clustering coefficient $c_{e_i}$ of a subgraph $k$-hops away from $e_i$. 
The regression loss is the MSE: 

$L_{\lcc} = \sum_{i=1}^k(f(e_i) - c_{e_i})_k^2$.

In general, nodes with a higher degree tend to have a higher \lcc, because they are more likely to have a large number of neighbors. 
Node degree is used to assess neighborhood quality in link prediction algorithms such as Common Neighbor~\cite{PhysRevE.64.025102}. 
Our ablation study shows that \iva and \lcc are useful when combined with other pretraining signals.

\section{Baselines}
\label{app:baselines}

In table~\ref{tab:link-prediction}, we compare our method against strong and recent KG Embedding methods. 
\boxe \cite{boxe} is a spatio-translational graph embedding model that uses logical rules (similar to paths) and that can support hierarchical inference patterns and higher-arity relations (knowledge hypergraphs or KHG). 
This is one of the rare methods that can be applied to both graphs and hypergraphs. \boxe achieves state-of-the-art results on \jf while remaining competitive on \fb and \wnrr. 
It is unclear however if \boxe is pretrainable\footnote{In table~\ref{tab:link-prediction}, we wrote ``N/A'' since we are not certain if pretraining is applicable.}. 
To our knowledge, \drgi \cite{drgi2021} is the only other KG completion method that was pretrained using only signals from the graph structure. 
Similarly to Deep Graph Infomax~\cite{velickovic2018deep}, \drgi is pretrained on artifacts of the graph structure by maximizing the mutual information between local and global graph representations. 
We were not able to ascertain at the moment if \drgi is applicable to KHGs.

Both \lpbert~\cite{lpbert2022} and \starbert~\cite{star2021} are Transformer-based Contextualized KG Embedding methods.
The two techniques are improvements over \kgbert~\cite{Yao2019KGBERTBF} that uses textual descriptions of entities and relations on top of a \bert model, which is pretrained on a large text corpus\footnote{The text pretraining data is external data and not self-contained to the KG data}. 
\lpbert also uses a multitask pretraining step by jointly training on three denoising tasks: random word masking in the descriptions, entity masking, and relation masking. 
\starbert combines \kgbert and TransE~\cite{transe2013}. 
\starbert contextualizes the translation function by embedding the head and relation descriptions with \bert. 
The authors claim that the technique is structure-aware since translation-based graph embedding approaches conduct structure learning by measuring spatial distance.
Note that KHG tuples in \jf can have up to 6 entities.
In most cases, the description of all entities in a tuple exceeds the standard \bert maximum sequence length of 1024.
For this reason, we were not able to apply \lpbert and \starbert to \jf.

\section{Experimental Set-Up}
\label{app:setup}


We use \fb \cite{freebase} (extracted from Wikidata, formerly Freebase) and \wnrr \cite{wn18rr}. 
We also test our method on a hypergraph link prediction task based off the \jf \cite{jf17k} dataset. 
Note that all datasets are heterogeneous graphs.
We see from the Table~\ref{tab:data} that we used datasets with a varying number of entities $\mathcal{E}$, number of relations $\mathcal{R}$, number of examples, density, and arity.

\begin{table}[H]
\small
\centering
\setlength{\tabcolsep}{2.5pt}
\begin{tabular}{@{} l| *7c @{}}
\multicolumn{1}{c|}{Dataset}& $|\mathcal{E}|$  & $|\mathcal{R}|$  & \#train  & \#valid & \#test & density & arity\\
\hline
 \fb & 14,541 & 237 & 272,115 & 17,535 & 20,466 & 18.7 & 2\\
 \wnrr & 40,943 & 11 & 86,835 & 3,034 & 3,134 & 2.1 & 2\\
 \hline
 \jf & 29,177 & 327 & 61,911 & 15,822 & 24,915 & 35.9 & $\ge$2\\
\end{tabular}
\caption{Dataset Statistics.}\label{tab:data}
\end{table}

The individual pretraining signals seem to lower precision (H@1) and increase recall (H@10) compared to \kgtrsf from scratch. 
Further, individual pretraining signals have a larger positive effect as graph density increases. 
Individual pretraining signals typically show MRR improvements for \jf, the densest of the KGs.

\paragraph{\textbf{Training and Evaluation}}

MRR is the Mean Reciprocal Rank and H@(1,3,10) are HIT@ measures all commonly used in link prediction \cite{Mohamed2019LossFI}.
For all tasks, we use the same autoregressive Transformer model that applies a transformation $f$ on the input and that is optimized with different loss functions.
Our Transformer model is a single monolithic architecture that uses a masking scheme similar to UniLM \cite{unilm2019}, allowing the model to play a variety of roles (encoder-only, decoder-only, or encoder-decoder) and tackle a variety of objectives (classification, regression or sequence generation). 
In our experiments, we have used $L=12$ Transformer layers, with $D=256$ hidden dimension, $A=12$ self-attention heads, an intermediate layer of size $2D$.
For path algorithms, we truncate our path sequences (max hops) to $l=4$, and, for neighborhood-based algorithms, we limit $k=3$.
The pretraining data generated from each dataset is applied to separate model instances. 
For each dataset, we save the weights of the pretrained model that performs best on the evaluation set of the link prediction task. 
During link prediction finetuning, we use a dropout rate $\rho \in \{0.1, 0.2, 0.3, 0.5\}$, a label smoothing rate $\zeta \in \{0.5, 0.6, 0.7, 0.8, 0.9\}$ and a learning rate of $\eta=5\time10^{-4}$.
In our multitask setting (\all), each task is sampled from the uniform distribution $\frac{1}{|D_t|}$, for dataset $D_t$ of a pretraining task~$t$. A batch may therefore contain multiple tasks. We weight the losses of tasks from $\tau = \{\spa, \ipa, \hn, \lcc, \iva \}$ according the available pretraining dataset size such that $L_{\all}= \sum_{t\in \tau}\alpha_{t}L_{t}$, where the $\alpha_{t} = |D_t|/\sum_{t'\in \tau}|D_t'|$.
For \hn, note that if $|E_{\mathcal{N}_k-1}|> 1024$, we clip the token sequence for a maximum length of 1024.
At inference, each entity in~$Q_r$ is masked and evaluated once. 
We use two evaluation metrics: HIT@$n$ and Mean Reciprocal Rank (MRR).

\section{Experiments}
\label{sec:exp}

We compare our results with several strong baselines described in Section~\ref{app:baselines}. 
Our results are presented in Table \ref{tab:link-prediction}. 
Our baseline model, \kgtrsf or ``CoKe'' \cite{coke}, is a Transformer model that is trained from scratch on link prediction. 
It is important to keep in mind the dataset properties in Table~\ref{tab:data} since they explain performance variations of the most recent techniques.
For example, \boxe is strongest for higher arity datasets such as \jf but lags behind on arity-2 KGs.
Similarly, \starbert has much better scores when evaluated on high density graphs such as \fb, 
while \lpbert is best for low density graphs such as \wnrr.
\kgtrsf \all however consistently performs well on three types of graphs.

Our new algorithm \ipa provides benefit over \kgtrsf across \emph{all} datasets. 
Further, we see that all H@ measures and MRR are typically higher for all pretraining signals compared to \kgtrsf.
Moreover, all pretraining signals (except \lcc) provide gains on the hypergraph \jf compared to \kgtrsf.
We see that using \all pretraining tasks jointly provides $5$\% relative MRR score increase over the unpretrained baseline, and often surpassing the competitive performances of state-of-the-art models.
In a multitask setting that combines all pretraining tasks, our method surpasses the latest and strong performing knowledge graph embedding methods on all metrics for \fb, on MRR and Hit@1 for \wnrr and on MRR and hit@10 for \jf (a knowledge hypergraph dataset).
The \all results are in-line with recent graph pretraining methods \cite{Hu_2020Strategies,gpt_gnn2020} that show that multitask pretraining performs better than any individual task.
Interestingly, our path-based algorithm \ipa provides the largest single task increase in performance over the unpretrained baseline.

Except for our \wnrr results, path-based pretraining surpasses neighborhood-based pretraining.
Compared to \spa, we hypothesize that \ipa provides a higher quality pretraining signal for two reasons: (1) paths are more diverse and (2) paths are semantically closer to a query $Q_r$. 
\spa produces paths that have often redundant first 2-hop relations on a k-hop path $\{P_{rel}\}_{r}$.
Further, all the paths have the same minimum length and often transit through high degree entities.
As seen in Section \ref{sec:shortest_path}, high degree entities do not necessarily provide the most meaningful reasoning chains; though the \ipa-based reasoning chains seem more semantically relevant. 
For example, for a masked query ``Spain'', ``form of government'' and $\mask{}$'', \ipa paths are [``military conflict/combatants'', ``international organization/member states''], [``adjoining relationship/adjoins'', ``continents/countries within''] or [``organization member/member of'', ``international organization/member states''].

\section{Analysis and Ablation}
\label{sec:ablation}
In this section, we discuss the choice and combination of tasks. An ablation study is presented in table~\ref{tab:ablation} that allows us to compare various multitask combinations. We first notice that given \spa, \ipa or \spa + \ipa relational path-based pretraining schemes, adding any other pretraining signal (\hn, \lcc or \iva) typically results in an improvement. \hn provides the largest MRR increase when combined with relational path-based signals, with \spa + \ipa + \hn providing a 0.06 improvement in MRR. \iva comes in as the second best signal to combine and \lcc only improving MRR slightly. 

\begin{table}[H]

\setlength\tabcolsep{4.5pt}
\begin{center}
\begin{tabular}{rrr|rrr|rrr}
\toprule
\rowcolor{cyan!16}\multicolumn{9}{c}{\it \textbf{Pretraining Signal Combination (MRR)}} \\
\multicolumn{2}{l|}{\spa} & 0.368 &  \multicolumn{2}{l|}{\ipa} & 0.372 &  \multicolumn{2}{l|}{\spa + \ipa} & 0.374 \\

\multicolumn{2}{c|}{+ \hn}  &  0.372 & \multicolumn{2}{c|}{+ \hn}  &  0.375 & \multicolumn{2}{c|}{+ \hn}  & \textbf{0.379} \\
\multicolumn{2}{c|}{+ \lcc} &  0.369 & \multicolumn{2}{c|}{+ \lcc} &  0.372 & \multicolumn{2}{c|}{+ \lcc} & 0.375 \\
\multicolumn{2}{c|}{+ \iva} &  0.370 & \multicolumn{2}{c|}{+ \iva} &  0.374 & \multicolumn{2}{c|}{+ \iva} & 0.376 \\

\bottomrule
\end{tabular}
\caption{\label{tab:ablation} Task combination ablation study on \fb using MRR. Adding individual signals to \spa, \ipa or \spa + \ipa pretraining schemes.}
\end{center}
\end{table}

\section{Conclusion}

We have investigated the effectiveness of five graph algorithmic pretraining schemes that do not rely on external data sources. 
On three downstream KG completion tasks, we found that \kgtrsf: 
(1) multitask pretraining results in performance increases, (2) generally, pretrained models exhibit better improvements in recall (H@10) rather than precision (H@1), and (3) the \ipa pretraining tasks work best.
Our study has important implications since it shows that using various graph structural signals that do not rely on external data can outperform strong baselines pretrained with external data.
A deeper study on graph topology and pretraining is still required (e.g., number of unique relation types, graph diameter). 
In future work, it would be interesting to investigate our single-datasets pretraining with entities and relations encoded with a BERT model.

\section*{Limitations.} Our technique has a few limitations. 
First, the pretraining signals that we used in this paper require a highly adaptive model. 
Out of the five pretraining schemes, two include the task of relational path sequence generation (\spa and \ipa) and another is a local clustering coefficient regression task (\lcc). 
Such tasks typically cannot be performed by most Graph Neural Networks or Graph Embedding methods. 
However, the versatility of our Transformer-based technique also means that our model can be pretrained on multiple modalities \cite{limoe2022,multigametrf2022}. 
For example, pretraining with text and KGs has already proven very powerful in language generation tasks \cite{agarwal-etal-2021-knowledge}. 
The point of our study was to show that we can surpass other methods such as \drgi and \lpbert (see  Section~\ref{app:baselines} for an overview of baselines) that use pretrained \bert models and entity descriptions. 
We suspect that enhancing our entity and relation representations with text will only make the model even stronger at KG completion. 
Finally, we notice that jointly training on \all pretraining tasks yields the best results. However, since it is possible that negative task interference occurs (a negative side effect of multitask learning \cite{negtrans2020,pilault2021conditionally}, a more throughout study of task combinations can help unlock even better performances for each specific dataset.

\clearpage


\bibliography{kr23}
\bibliographystyle{kr}

\clearpage

\end{document}